\documentclass[wcp]{jmlr}
\usepackage{longtable}
\usepackage{multirow}
\usepackage{microtype}
\usepackage{booktabs}
\usepackage{lineno}
\makeatletter
\let\Ginclude@graphics\@org@Ginclude@graphics 
\makeatother

\jmlrvolume{222}
\jmlryear{2023}
\jmlrworkshop{ACML 2023}

\title[Meta-forests: Domain generalization on random forests with meta-learning]{Meta-forests: Domain generalization on random forests with meta-learning}

\author{\Name{Yuyang Sun} \Email{yuyang.1.sun@kcl.ac.uk}\\
 \Name{Panagiotis Kosmas} \Email{panagiotis.kosmas@kcl.ac.uk}\\
\addr Department of Engineering, King’s College London, UK}

\editors{Berrin Yan{\i}ko\u{g}lu and Wray Buntine}

\begin{document}

\maketitle

\begin{abstract}
Domain generalization is a popular machine learning technique that enables models to perform well on the unseen target domain, by learning from multiple source domains. Domain generalization is useful in cases where data is limited, difficult, or expensive to collect, such as in object recognition and biomedicine. In this paper, we propose a novel domain generalization algorithm called "meta-forests", which builds upon the basic random forests model by incorporating the meta-learning strategy and maximum mean discrepancy measure. The aim of meta-forests is to enhance the generalization ability of classifiers by reducing the correlation among trees and increasing their strength. More specifically, meta-forests conducts meta-learning optimization during each meta-task, while also utilizing the maximum mean discrepancy as a regularization term to penalize poor generalization performance in the meta-test process. To evaluate the effectiveness of our algorithm, we test it on two publicly object recognition datasets and a glucose monitoring dataset that we have used in a previous study. Our results show that meta-forests outperforms state-of-the-art approaches in terms of generalization performance on both object recognition and glucose monitoring datasets.
\end{abstract}
\begin{keywords}
Domain generalization, Meta-learning, Random forests.
\end{keywords}

\section{Introduction}
Machine learning (ML) has made significant advances in various fields, including computer vision, natural language processing, and biomedicine, resulting in highly accurate models. Nevertheless, collecting a large dataset for training these models is often difficult and time-consuming, especially in areas like biomedicine where data collection can be expensive. Furthermore, the collected datasets exhibit variations in their distributions due to differences in data collection scenarios, devices used, and the subjects involved. These variations, referred to as domain shifts, pose a significant challenge to traditional ML approaches that assume identical and independent data distributions. Domain shifts can cause a decrease in model performance when trained on one domain but deployed on another, as the underlying data distributions differ. In such scenarios, traditional ML models struggle to generalize effectively, leading to poor performance and reduced accuracy. 

Non-invasive glucose sensing is a crucial biomedical application that can benefit from ML techniques. However, the collection of training data for this application, particularly from diabetes patients using devices that are still under development, is challenging in terms of time, cost, and data availability. Clinical trials, which are necessary for gathering relevant data, can be time-consuming and expensive, making it difficult to acquire a large dataset suitable for training traditional ML methods, such as neural networks. Furthermore, the variability among subjects introduces differences in the data distribution, further complicating the application of standard ML techniques. Consequently, there is a need to develop new Domain Generalization (DG) algorithms for biomedical applications, which can enhance the generalization ability of ML models when applied to limited and not identical image and signal datasets.

DG is a ML technique that enables learning knowledge from multiple related but different distribution domains and the ability to generalize well on an unseen test domain. In contrast to related topics such as domain adaptation and transfer learning, DG emphasizes that the dataset comprises samples from different distribution domains, without access to the test domain during training, while ensuring that the training and test labels are in the same label space. Although several DG algorithms have been proposed for object recognition tasks, such as~\cite{cite_02, cite_03, cite_07, cite_08, cite_09, cite_11, cite_16, cite_17}, the application in the field of biomedicine is limited due to the scarcity and diversity of biomedical data compared to other fields.

Object recognition tasks benefit from large-scale publicly available datasets, such as ImageNet by~\cite{cite_30}, DomainNet by~\cite{cite_31}, and Open Images by~\cite{cite_32}, which provide a large amount of labeled data for training and evaluation. In contrast, biomedical datasets, especially those related to clinical applications, face challenges about privacy concerns, ethical considerations, and the complexities of collecting data from human subjects. Besides, biomedical data includes text, numerical, and signal data in addition to images. Directly applying DG algorithms designed for image datasets may not yield satisfactory generalization performance when applied to other data types.

To address these challenges, we propose a DG algorithm specifically designed for the area of non-invasive glucose detection, a crucial research topic in biomedicine. To this end, we use a previously collected blood glucose monitoring dataset in~\cite{cite_add04}, which includes various types of data such as radio frequency signals, near-infrared (NIR) signals, temperature, and pressure data to evaluate the performance of our proposed algorithm. Furthermore, we compare the performance of our algorithm with other state-of-the-art DG algorithms by evaluating them on two publicly available object recognition datasets.

Unlike other DG algorithms that rely on deep neural networks, our algorithm is based on an ensemble model, random forests. Compared to other deep models, the random forest algorithm requires fewer data for training, is less prone to over-fitting, and has high interpretability. Typically, the random forests model can provide feature attribution and feature importance for analysis as discussed in~\cite{cite_add06}. Additionally, the generalization error bound of the random forests is bounded by two factors: the strength, and the correlations of the set of single classifiers (trees) by~\cite{cite_04}. To achieve generalization, it is crucial to ensure the forests have high strength and low correlations. However, these two factors are in conflict with each other, similar to the well-known bias-variance trade-off in ML. When trees in the forests have high strength, they tend to have similar shapes due to the bootstrap mechanism, leading to high correlations. Thus, our proposed algorithm aims to address the trade-off between high strength and low correlation in the forests to achieve good generalization on the unseen domains.

In the context of our proposed DG algorithm, the term 'domain' refers to data with the same distribution within a dataset. It is worth noting that the traditional ML approach uses the terms 'training' and 'test' sets to refer to the data used for model training and evaluation, respectively, while the terms 'source' and 'target' domains are used in our approach. The primary objective of DG optimization is to minimize the risk between source and target domains resulting from domain shifts. In summary, our proposed algorithm aims to enhance the model's ability to generalize well on previously unseen domains by minimizing the generalization error caused by domain shifts.

In our study, we extend the DG approach with meta-learning to construct a series of meta-tasks that include multiple meta-train and meta-test sets. In each iteration, the model is trained to maximize accuracy on the meta-train set to enhance its strength. We also introduce maximum mean discrepancy (MMD) as a measure of the distances among different domains. MMD is used as a regularization term to describe the distances between the meta-train and meta-test sets and penalize the large domain shift by the weights update function. Through several iterations of meta-learning and weights update functions, a generalized random forest model is generated on the generalized distribution space based on different source domains, and the model also generalizes well on the unseen target domain. To reduce correlations among trees while maintaining the strength of classifiers, we employ several approaches such as hyper-parameter setting and introducing randomness. These approaches improve the diversity of the forest model and increase the generalization ability of the model and prevent over-fitting to the source domains.

We summarize the contributions of our proposed work as follows: Firstly, we propose a DG algorithm, meta-forests, that enhances the generalization ability of the random forests algorithm. Secondly, we address the trade-off between the strength and correlation of the basic random forests by introducing the meta-learning strategy, MMD penalized term, hyper-parameter setting, and randomness. Lastly, we demonstrate the efficacy of our proposed algorithm on various datasets, including a previously collected blood glucose dataset and two public object recognition datasets.

\section{Preliminaries}
\subsection{Domain Generalization}
DG aims to minimize the error of a model on unseen domains by training it on several different distribution domains. Mathematically, DG can be described as follows: Let $X$ denote the input space and $Y$ denote the output space. We have a set of M source domains denoted by ${D^{s}= (X_{s}^{i},Y_{s}^{i})_{i=1}^{M}}$,
where ${X_{s}^{i}}$ and ${Y_{s}^{i}}$ are the input and output spaces of the i-th source domain, respectively. Similarly, we have a target domain denoted by ${D^t = (X_t, Y_t)}$, where ${X_t}$ and ${Y_t}$ are the input and output spaces of the target domain. For DG, we aim to learn a model ${f:X\to Y}$ which is trained on the source domains ${D^{s}}$ and generalizes well to the target domain ${D^{t}}$, despite not having any labeled data from the target domain. 

The goal of DG optimization is to minimize the generalization error on unseen target domains. As reported in previous studies by~\cite{cite_01} and~\cite{cite_07}, the generalization error of DG is bounded by the source risks among source domains and the differences between the source domains and the target domain. The source risk among source domains is primarily caused by the distribution differences among them, which is challenging to reduce. Therefore, the main objective of DG is to reduce the distribution distance between the source domains and the target domain.

\cite{cite_02} proposed an algorithm that utilizes adversarial data augmentation to iteratively generate data and improve data quality.~\cite{cite_03} presented an algorithm that aligns source distributions to a known prior Laplace distribution, thereby enhancing model generalization.~\cite{cite_08} divided domains into multiple tasks and employed a series of iterative meta-learning strategies to achieve DG.~\cite{cite_09} designed models to learn specific source knowledge and combined them to obtain the final prediction. These algorithms employ data manipulation, representation learning, or learning strategy adaptation to address domain shifts resulting from disparate distributions.

Our approach integrates the feature alignment and ensemble learning approaches to effectively merge the domain invariant knowledge learning and target domain generalization strategies. Specifically, we modify the training strategy of the ensemble model (i.e., random forests) using meta-learning and introduce MMD to align the feature distributions between source domains and the target domain. Due to the smaller amount of training data required by random forests compared to deep models, our proposed algorithm, meta-forests, achieves state-of-the-art accuracy while utilizing a smaller amount of data.

\subsection{Random Forests}

Random forests is an ensemble model that combines multiple weak tree classifiers to vote unweighted decisions. The training process of random forests involves a bootstrap mechanism, where a portion of the data is randomly selected for training and constructing the forests at each iteration. The ensemble structure and utilization of the bootstrap mechanism in random forests contribute to its high accuracy while effectively reducing the risk of over-fitting according to the work by~\cite{cite_10}. However, the correlations among the forests are also high due to the training data having the same distribution and the forests having high strength.

Equation~\eqref{eqn:1} demonstrates the generalization error bound for random forests, which is bounded by the strength and correlation of the forests. Chebyshev's Inequality can be used to derive this error bound, with more details in~\cite{cite_04}.

\begin{equation}
\label{eqn:1}
    PE^{*} \leq \frac{\bar{\rho }\left ( 1-s^{2} \right )}{s} \\
\end{equation}
In Equation~\eqref{eqn:1}, $PE^{*}$ is the generalized error of random forests classifiers, $\bar{\rho }$ is the mean value of correlations among classifiers in the random forests, and $s$ is the strength of the set of classifiers.

There is limited research focusing on generalized random forests.~\cite{cite_11} proposed a method for modifying tree representations using triplet sampling to increase correlation while maintaining strength. Similarly,~\cite{cite_06} pruned some nodes in the forests that do not affect their prediction strength to achieve generalization in transfer learning settings. These approaches aim to improve the generalization ability of random forests by changing the structure of the trees. In contrast, our proposed method maintains the generated forest structure and instead introduces randomness during the generation process. Furthermore, we incorporate a meta-learning strategy to learn a set of weights for the trees in the forest, which modifies the unweighted voting mechanism and enhances the model's ability to generalize to unseen domains.

\subsection{Mean Maximum Discrepancy}

MMD is a common metric to measure the distance between different distribution domains in the DG problem. MMD calculates the difference between the average feature representations of two distributions in Hilbert space, providing a measure of the distance between the two distributions. Specifically, given two probability distributions $P$ and $Q$ over sets $X$ and $Y$ in the Hilbert space $\mathcal{H}$, the MMD between them is calculated using Equation~\eqref{eqn:2}:

\begin{equation}
\label{eqn:2}
MMD(P,Q) = \sup_{f\in \mathcal{H}, \left \| f \right \|_{\mathcal{H}}\leq 1}E_{X\sim P}\left [f (X) \right ]-E_{Y\sim Q}\left [f(Y) \right ]
\end{equation}
where $E_P, E_Q$ are the expectations of distributions $P, Q$, $f(.)$ is the mapping function that maps the input space to the Hilbert space $\mathcal{H}$, and the norm of $f(.)$ in $\mathcal{H}$ space cannot be greater than $1$. The calculation of MMD enables us to quantify the distance between the two distributions.

Previous works on DG have widely used MMD as a metric for measuring the distance between distributions. For example,~\cite{cite_03} introduced MMD to align distributions to a Laplace distribution using an adversarial learning framework. Similarly,~\cite{cite_12} proposed a canonical correlation analysis model with MMD measure as a regularization term for measuring domain distances, with the aim of achieving domain-invariant feature learning. By introducing MMD as a regularization term, the distance between distributions of different domains can be reduced, thereby contributing to the goal of achieving generalization in DG tasks.

In our work, we adopt a similar approach by using MMD to measure domain distances and introducing it into the process of updating the weights of the random forests. As an ensemble model, random forests can benefit from assigning different weights to different distributions, in order to achieve a generalized distribution layout that performs well in unseen domains. By leveraging MMD in this way, we aim to improve the generalization ability of the random forests in DG tasks.

\section{Meta-forests}

In this section, we present our proposed DG algorithm, meta-forests, which is based on basic random forests. Our approach aims to increase the strength of random forests by leveraging two mechanisms: meta-learning and domain alignment. We also describe how we decrease the correlations among trees in the random forests, without reducing their strengths.

\begin{figure*}[htb!]
  \centering
  \centerline{\includegraphics[width = 0.8\linewidth,height = 3in]{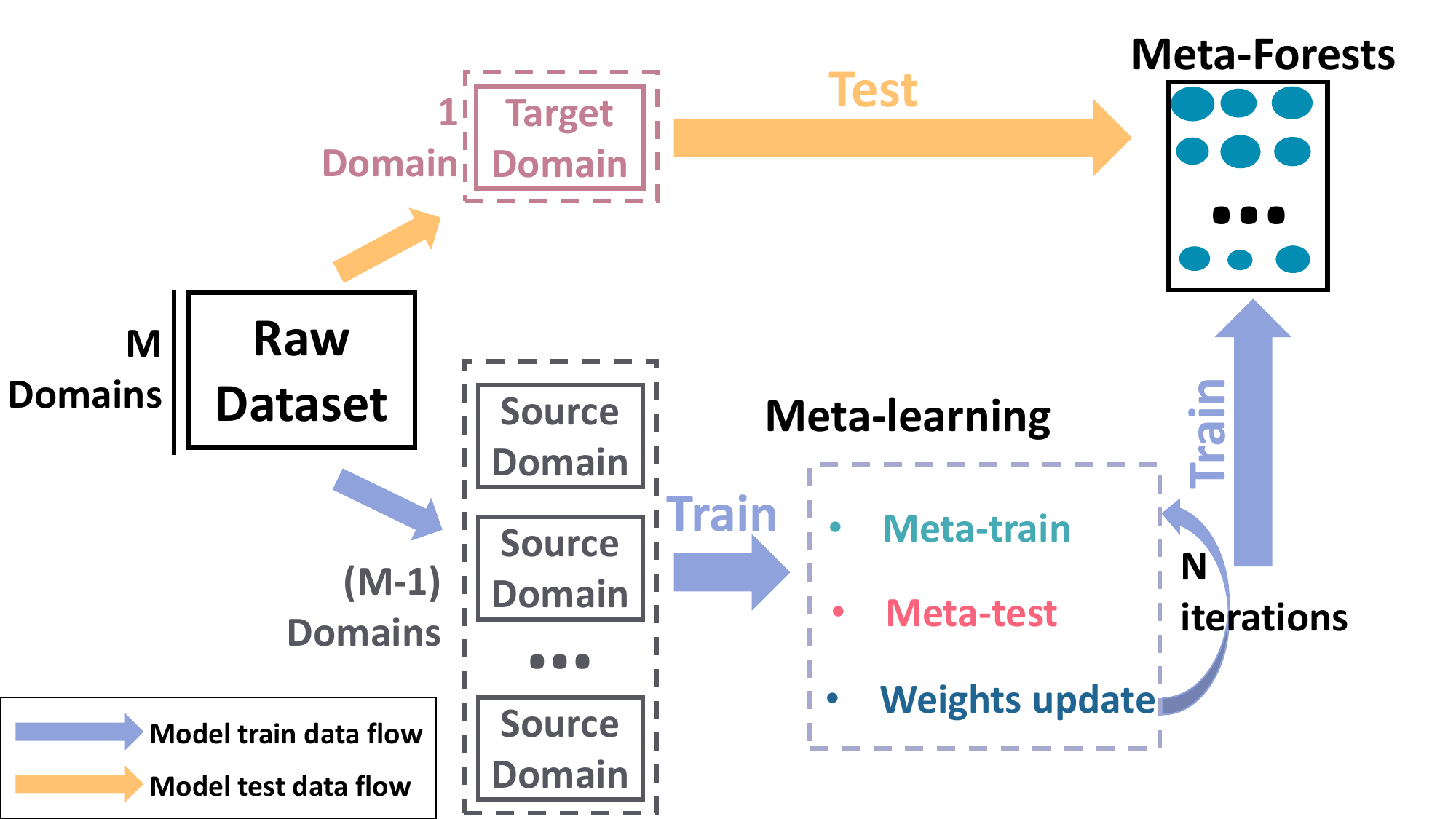}}
  \caption{An overview of our proposed domain generalization algorithm meta-forests.}
\label{fig:fig_label_0}
\end{figure*}

The overall flowchart of the meta-forests algorithm is depicted in Figure~\ref{fig:fig_label_0}. Firstly, we split the raw dataset with M domains into one target domain and (M-1) source domains according to domain distribution differences.  Next, we use the data from (M-1) source domains to train meta-forests for N iterations using the meta-learning strategy. The target domain remains unseen during the training process and is only used for testing the final model's performance.

\subsection{Meta-learning for Meta-forests}

\begin{figure*}[htb!]
  \centering
  \centerline{\includegraphics[width = 0.9\linewidth,height = 2.95in]{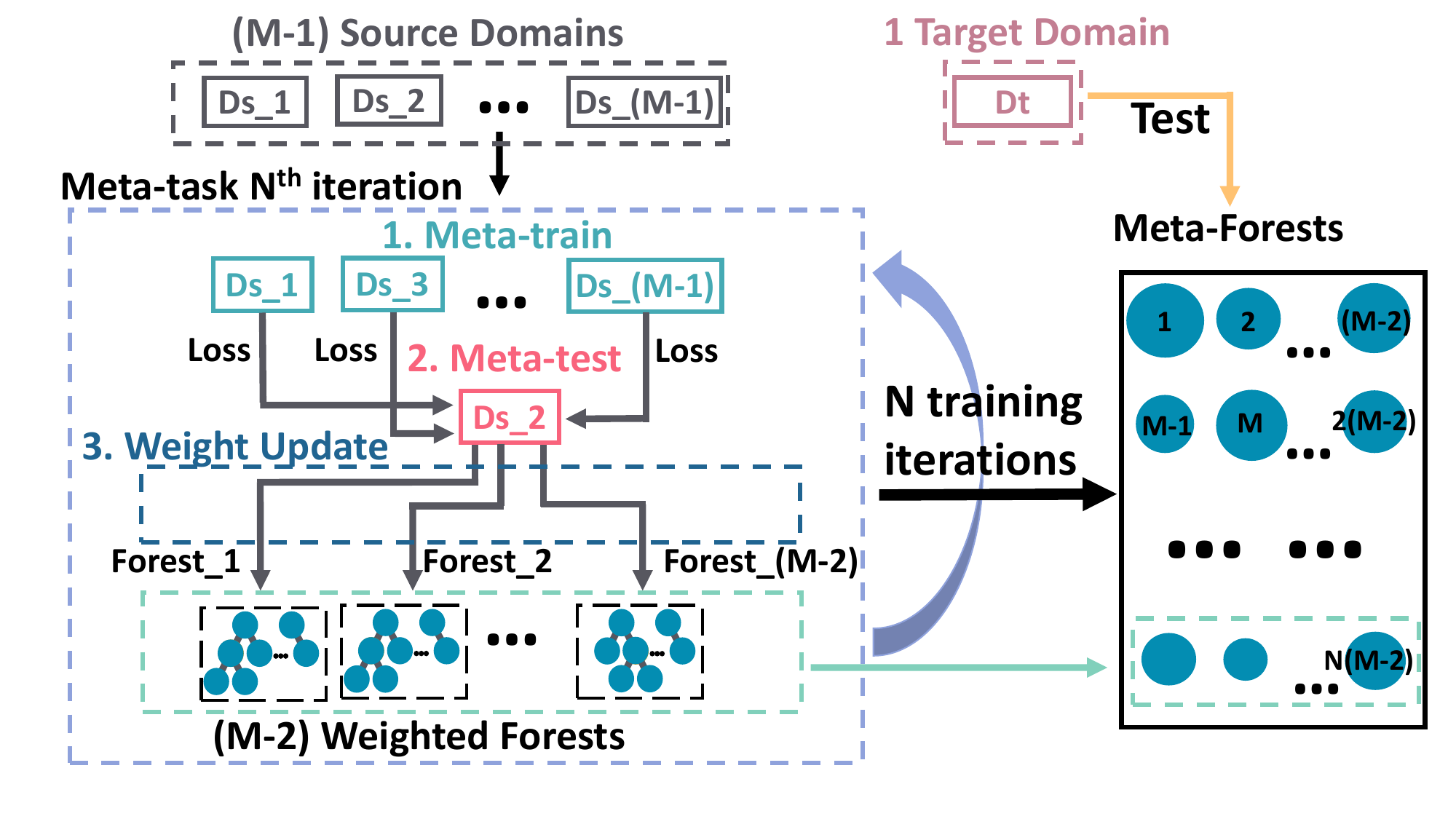}}
  \caption{Structure of meta-forests domain generalization algorithm, which are details about the 'Meta-learning' box in Figure~\ref{fig:fig_label_0}. The source domains are denoted as $D_s$, and the target domain is denoted as $D_t$. At each meta-task training iteration, one of the source domains is randomly selected as the meta-test set, while the remaining domains serve as the meta-train set. For instance, at the $N^{th}$ iteration, we select $D_s\_2$ as the meta-test set, while in previous iterations, it was randomly selected from the other domains. Within each meta-task iteration, meta-learning and weights update are performed, resulting in (M-2) weighted forests. The meta-task is repeated N times, and all N(M-2) weighted forests generated are collected to construct the meta-forests model. Finally, the produced meta-forests model is evaluated on the target domain $D_t$. }
\label{fig:fig_label_1}
\end{figure*}

Meta-learning is a type of learning strategy that aims at learning the algorithm itself by the previous tasks, a concept also known as 'learning-to-learn'. Traditional ML algorithms have poor generalization performance in the DG problem due to differences in feature space distribution across domains, which can lead to low accuracy/strength in unseen domains. To address these issues, we introduce meta-learning as a learning strategy to enhance the generalization performance of the ensemble model, random forests. At each iteration of the meta-learning process, we introduce a penalty term based on MMD to align feature space distributions across source domains and determine the weights of the trees in a particular distribution subset. The approach aims to find a distribution layout that approximates the target distribution. We summarize our proposed approach in \textbf{Algorithm 1} and illustrate details of the process of meta-learning strategy in Figure~\ref{fig:fig_label_1}.

\begin{algorithm}[htb!]
\hrulefill
\vspace{-.1cm}

\textbf{Algorithm 1} Meta-learning for Meta-forests

\vspace{-.3cm}
\hrulefill

\KwIn{
Source domain $D_{source}$, 
The number of source domain $M-1$, The number of iterations $N$, Weight update functions $\mathcal{W(.)}$.}

\KwOut{Weighted meta-forests set ${F_{1,(M-2)}, F_{2,(M-2)}, ..., F_{N,(M-2)}}$ and their respective weights ${W_{1,(M-2)}, W_{2,(M-2)}, ..., W_{N,(M-2)}}$.}

\textbf{Initialize} 
 $\alpha$, $\beta$ and $W_{0} = \frac{1}{M-2}$;

\For{$i=1$ \KwTo $N$}{
Randomly select $1$ domain $D_{meta\_test}$ from $D_{source}$ ; 

Let $D_{meta\_train}$ be the remaining $(M-2)$ source domains in $D_{source}$ ;

\For{$j=1$ \KwTo $M-2$}{
  Train random forest model $F_{j}$ on $D_{j}\in D_{meta\_train}$ with weights $W_{i-1,j}$\;

}

Calculate $W_{mmd}, W_{accuracy} $ by $\mathcal{W}(F_{i,j})$ using Eq.\eqref{eqn:3};

Update the weights $W_{i,j}$ $= W_{i-1,j} \times e^{\alpha W_{mmd}}\times \log^{\beta W_{accuracy}}\nonumber$;

Normalize the produced weights $W_{i,j}$ of $F_{i,j}$; 

}

\vspace{-.2cm}
\hrulefill
\end{algorithm}

In our proposed approach, we assume there are $(M-1)$ source domains and $1$ target domain. Although the source and target domains have the same label space, their feature spaces have different distributions. The source domains are divided into two sets, namely the meta-train and meta-test sets. Each meta-task consists of a pair of these sets. At each meta-task iteration, we randomly select one source domain as the meta-test set, and the remaining source domains are as the meta-train set. We use a portion of data from each domain in the meta-train set to train the random forests models individually, with initialized or updated weights. During the meta-test phase, we use a portion of the previously selected meta-test set to calculate the weights of the newly generated random forests. The function for the weights update function is shown in Equation~\eqref{eqn:3}.

\begin{eqnarray}
\label{eqn:3}
W_{i,j} = W_{i-1,j} \times e^{\alpha W_{mmd}}\times \log^{\beta W_{accuracy}}\nonumber\\
W_{mmd} = mmd_{i,j}-\frac{\sum_{x = 0}^{i-1}\sum_{y=0}^{j-1}mmd_{x,y}}{(i-1)*(j-1)}\nonumber\\
W_{accuracy} = e^{{\frac{\sum_{n=1}^{S}I(y_n = \hat{y}_n)}{S}}}-e^{\frac{1}{C}}
\end{eqnarray}

Here, $i$ (where $i=1,2,...,N$) represents the iteration number, and $j$ (where $j=1,2,...,M-2$) represents the meta-train domain number. The coefficients $\alpha$ and $\beta$ (where $\alpha <0, \beta>0$) correspond to the MMD weight and accuracy weight, respectively. The $W_{mmd}$ is a decreasing function, which means that as the MMD distance between domains increases, its weight becomes smaller, and vice versa. On the other hand, the $W_{accuracy}$ is an increasing function, which means that as the model's accuracy on the meta-test set improves, its weight becomes larger, and vice versa. $S$ is the total number of samples in the meta-test set, $y_n$ is the true label of the $n$-th sample, $\hat{y}_n$ is the predicted label of the $n$-th sample, $I(\cdot)$ is the indicator function which returns 1 if its argument is true and 0 otherwise, and $C$ is the number of classifications.

The weight update function of our proposed meta-learning strategy comprises two components: accuracy weight $W_{accuracy}$ and MMD weight $W_{mmd}$. The accuracy weight reflects the performance of a newly generative random forest model on the meta-test distribution. The MMD weight represents the distance between the meta-test distribution and the other source domain distribution. In the forests model, the function incorporates penalization if the accuracy is below the average prediction model accuracy or the distribution distance exceeds the average MMD distance in the current iteration. The objective of the weights update function is to assign more weight to models that exhibit high accuracies and low domain shifts on the meta-test set. When each iteration is finished, all the weights of newly generated and old trees should be normalized. The iteration process continues until the preset iteration value of $N$ is reached, where $M<<N$. Upon completion of the meta-learning process, we obtain a weighted meta-forests set consisting of $N*(M-2)$ random forest models and their respective calculated weights, along with an unseen target domain dataset. The weighted model has good accuracy on the combination layout of all source domain distributions and is generalized well on the unseen target domain dataset. The final prediction results on the target domain are the weighted averages based on produced forests and the updated weights matrix. Meanwhile, we adopt some approaches to reduce the correlations among classifiers except the meta-learning strategy, which will be discussed in the following Section 3.2.
\subsection{Reducing Correlation of Meta-forests}

The main reason for high correlations between trees in the random forest is the low randomness while constructing trees from the same distribution. Specifically, the generated trees are not built ensembles using different subsets of features and training data. If the algorithm is not provided with sufficient randomness, such as by using a small number of trees or using the same random state seed for each tree, the trees end up becoming more similar to each other. This can lead to over-fitting and a reduction in the diversity of the ensemble, which can ultimately result in poorer performance on new, unseen data. To reduce this risk, we take two approaches to ensure there is enough randomness while generating the trees of random forests.

Firstly, we change a series of hyper-parameter settings while training the random forests. To ensure sufficient randomness, it is important to build a large number of trees in a forest and select different random state seeds for each tree, ensuring that each tree is constructed using a different random subset of features and data. Besides, the training data is randomly sampled with replacement and the size of sampling is small to create various subsets of the same distribution data for each tree. Additionally, the maximum depth of each random forests is also constrained to prevent the model split training distribution samples too small that over-fitting.

Secondly, in addition to adjusting hyper-parameter settings, we further modify the model selection method of the basic random forests algorithm. Specifically, we incorporate a feature subsampling technique to reduce the correlation between trees in the ensemble. Instead of considering all available features at each split in each tree, we randomly select a subset of features. Importantly, the features used for splitting in one iteration are masked for the subsequent iteration within the same distribution. This deliberate alteration of the feature selection mechanism affects the individual tree's strength, but it serves the purpose of reducing over-fitting by preventing the trees from excessively capturing noise in the data. Moreover, this approach enhances the diversity of the ensemble by ensuring that each tree is constructed using a distinct subset of features, while also reducing the computational complexity of the algorithm by minimizing the number of features considered at each split. In the subsequent sections, we demonstrate two experiments to validate the proposed algorithm's generalization performance on unseen domains.

\section{Experiments}
In this section, we conduct experiments on the blood glucose monitoring dataset presented in~\cite{cite_add04} to show the benefits of the random forests-based model on handling subjects' variance in the biomedicine dataset. Besides, we test our proposed meta-forests and other state-of-art DG models on two public datasets, VLCS by~\cite{cite_15} and PACS by~\cite{cite_33} multi-domain datasets for evaluating the models' generalized abilities on object recognition. 

\subsection{Datasets and Baseline Domain Generalization Models}

The glucose monitoring dataset is a non-invasive glucose monitoring dataset presented in~\cite{cite_add04} from real subjects. VLCS and PACS datasets are popular public object recognition datasets designed for cross-domain recognition tasks. Table \ref{tab:my-table_1} lists the details of those datasets.

\begin{table}[htb!]
\centering
\caption{Statistics of glucose monitoring and object recognition datasets.}
\begin{tabular}[c]{cccc}
\toprule
Dataset           & \#Sample & \#Class & \#Domain \\ \midrule
Glucose monitoring &  8716       & 3       & $P_{i}(i=1, 2, ...,5)$         \\
VOC2007           & 3376     &     5    & V       \\ 
LabelMe           & 2656     &    5     & L       \\ 
SUN09             &  3282        &    5     & S       \\ 
Caltech-101       &  1415       &    5     & C       \\ 
VLCS              &10729     & 5       & V, L, C, S \\ 
PACS & 9991     & 7      & P, A, C, S \\ 
\bottomrule
\end{tabular}
\label{tab:my-table_1}
\end{table}

The glucose monitoring dataset contains data from four sensors: millimeter-wave, NIR, temperature, and pressure sensors. The dataset consists of a total of 8,716 samples from 5 subjects ($P_{i}, i =1, 2, .., 5$), who were subjected to an intravenous glucose test. Due to the variance in physical and physiological factors among subjects, different subjects are set as different domains in the dataset, which have domain shifts. In each domain, a single sample contains 19 features, which include 13 $S_{21}$ transmission coefficient parameters of 13 different frequencies of millimeter-wave signals, 4 transmittances of 2 different wavelengths of NIR signals from 2 adjacent collection frames, 1 skin temperature, and 1 skin pressure. These features have been proven to reflect human blood glucose changes or affect blood glucose changes according to the research by~\cite{cite_19, cite_20, cite_21, cite_22, cite_23, cite_24}. The blood glucose measurements in the dataset are set as 3 classes based on glucose tolerance test results: Normal ($<$ 7.8 mmol/l), Prediabetes (7.8 - 11.1 mmol/l), and Diabetes ($>$11.1 mmol/l) by~\cite{cite_28}.

The VLCS dataset is a public object recognition dataset designed for cross-domain recognition tasks. It contains data from four domains: VOC2007(V), LabelMe(L), SUN09(S), and Caltech-101(C). The dataset has a total of 10,729 image samples ($224*224$ pixels) of 5 common classes: 'bird', 'car', 'chair', 'dog', and 'person'. It is worth noting that the data used in the VLCS dataset are from four large standard object recognition datasets, and the samples and classes we describe here are overlapping classes of the large object recognition dataset, which are commonly used in evaluating DG models. The PACS dataset is another public object recognition dataset designed for cross-domain recognition tasks. It consists of data from four domains: Photo(P), Art\_painting(A), Cartoon(C), and Sketch(S). The dataset contains a total of 9,991 image samples ($224*224$ pixels) of 7 classes: 'dog', 'elephant', 'giraffe', 'guitar', 'horse', 'house', and 'person'.

We compare our proposed meta-forests with basic models and DG models proposed in recent years. The basic models we evaluate on the glucose monitoring dataset are SVM by~\cite{cite_25}, random forests by~\cite{cite_04}, ExtraTrees by~\cite{cite_add01}, AdaBoost by~\cite{cite_add02}, XGBoost by~\cite{cite_add03}, and ResNet-18 by~\cite{cite_26}. Compared with those basic models, meta-forests showed improved generalized ability. The state-of-art models we examine include CCSA by~\cite{cite_27}, MMD-AAE by~\cite{cite_03}, D-MTAE by~\cite{cite_16}, GCFN by~\cite{cite_11}, Meta-reg by~\cite{cite_17}, and MLDG by~\cite{cite_08}. Meta-forests performs better than all of these state-of-art models on their designed object recognition area and requires smaller data sizes. In Section 4.2, we test meta-forests and other basic models on the collected glucose monitoring dataset to show it is more generalized over basic models. In Section 4.3, we demonstrate the experiments we conducted on the public object recognition dataset with meta-forests and previously mentioned DG models. These experiments provide a way to compare meta-forests generalized ability with state-of-art works.

\subsection{Experiments on Glucose Monitoring Dataset}

In our study, we conducted a series of experiments utilizing various machine learning models, including SVM, basic Random Forest, ExtraTrees, AdaBoost, XGBoost, and ResNet-18 neural networks. These experiments were based on the glucose monitoring dataset, and the data preprocessing process was consistent with the previous study~\cite{cite_add04}. The training approach of those models is different and summarized as follows:

\textbf{SVM}, \textbf{random forests}, \textbf{ExtraTrees}, \textbf{AdaBoost}, \textbf{XGBoost} and \textbf{ResNet-18}: In each round of experiments, one subject's domain of the dataset is selected as the target domain, and the remaining other domains are selected as source domains. For every model, we conduct $20$ experiments by using source domains as training data and the target domain as test data.

\textbf{Meta-forests}: The source and target domains split are the same as previous models' settings, but the source domains will split into meta-train/test sets further.  As algorithm 1 demonstrates, in every single meta-task iteration, we randomly select one subject domain from source domains as meta-test set and the remaining domains as meta-train set. Furthermore, 30 \% meta-train/test set data is selected as meta-learning training strategy data for a single meta-task iteration.

Table~\ref{tab:my-table_9} summarizes the average accuracy results of those $20$ experiments.

\begin{table}[htb!]
\centering
\caption{Average prediction classification accuracy of models on glucose monitoring dataset using subject-specific target domains. The table presents the average prediction classification accuracy (\%) of basic models and (accuracy (\%) $\pm$ standard deviation) of meta-forests when selecting a specific subject's data (denoted as $P_{i}$, where $i=1, 2, 3, 4, 5$) as the target domain, while using the remaining subjects' data as the source domain.}
\begin{tabular}{cccccc}
\toprule
Model          & $P_{1}$ & $P_{2}$ & $P_{3}$ & $P_{4}$ & $P_{5}$  \\ 

\midrule
SVM            & 39.2  & 53.1  & 37.6  & 40.9  & 39.5              \\
Random forests & 38.5  & 63.2  & 40.5  &  47.2 & 44.5             \\
ExtraTrees      & 42.6  & 48.7  & 40.3  & 50.6  & 45.8  \\
AdaBoost      & 39.2  & 51.8  & 41.1  & 45.7  & 40.3  \\
XGBoost      & 37,7  & 63.3  & 43.2  & 52.3  & 41.4  \\
ResNet-18      & 18.3  & 40.2  & 35.4  & 34.2  & 24.4

\\\midrule
\textit{\textbf{Meta-forests}}     & $45.4\pm0.41$ & $68.9\pm0.27$  & $59.8\pm0.31$  &  $78.4\pm0.28$ & $58.7\pm0.18$           \\ \bottomrule         
\end{tabular}
\label{tab:my-table_9}
\end{table}

Based on the results presented in Table~\ref{tab:my-table_9}, it can be observed that SVM and forest-based models perform better than ResNet-18 on the glucose monitoring dataset due to the dataset's small size of features. Deep neural networks require large amounts of data to perform well, and with a small number of features, they do not have enough information to effectively train a deep neural network. In addition, as~\cite{cite_29} reported, the relationships among $S_{21}$ parameters, NIR transmittance, and human blood glucose are non-linear. Therefore, the forest-based model is better than SVM on the glucose monitoring dataset.

Subsequently, a performance comparison was conducted between random forests and meta-forests to demonstrate the improved generalization ability of the meta-forest. Meta-forests is designed specifically to enhance the generalization performance of random forests, which is bounded by its strength and correlation, as presented in formula 1. The hyper-parameters of both random forests and meta-forests were set to the same values in each round of the experiment to evaluate the generalization improvement. The results in Table~\ref{tab:my-table_9} demonstrate that meta-forests improves the accuracy of the model for all available subject domain splits.

Moreover, experiments were conducted to explore the benefits of reducing correlations among trees model, as presented in Section 3.2, to improve the generalization of the models. Table~\ref{tab:my-table_5} presents the comparison of the results of random forests and meta-forests with approaches that reduce correlations among trees.
Based on the comparison experiments presented in Table~\ref{tab:my-table_5}, show that the approaches we applied to reduce correlations among trees can enhance the accuracy of both random forests and meta-forests on the unseen domain. 

\begin{table}[htb!]
\centering
\caption{Comparison of random forests and meta-forests with approaches reducing correlations among trees.}
\begin{tabular}{ccccccc}
\hline
\multirow{2}{*}{Model} & \multicolumn{3}{c}{Sample sampling} & \multicolumn{3}{c}{Max\_depth} 
\\ \cline{2-7} 

                       & 10\%       & 20\%       & 30\%      & 3        & 5        & 10       \\ \midrule
Random forests         &  39.2          &    41.8        &    40.3       &  32.4        &  45.1        &  29.8        \\
\textit{\textbf{Meta-forests}}          &   43.2         &    61.7        &     54.1      &  35.5        &   60.9       &   41.6       \\ \bottomrule
\end{tabular}

\vspace{.3cm}

\begin{tabular}{ccccc}
\hline
\multirow{2}{*}{Model} & \multicolumn{2}{c}{Random seeds} & \multicolumn{2}{c}{Feature sub-sampling} \\ \cline{2-5} 
                       & T               & F              & T                   & F                  \\ \midrule
Random forests         &     44.5            &     25.3           &  43.6                   &     39.7               \\
\textit{\textbf{Meta-forests}}           &     61.5            &     35.8           &  60.2                   &      58.6              \\ \bottomrule
\end{tabular}
\label{tab:my-table_5}
\end{table}

However, with the sampling ratio of samples increasing, more data is utilized for training forests within each meta-task iteration. This leads to a higher correlation among trees belonging to the same domain, consequently resulting in poorer generalization performance. Additionally, if limited data is used for one meta-task training, the model prediction exhibits poor accuracy due to inadequate learning from the data. Furthermore, the hyper-parameter max\_depth  bounds the maximum depth of random forests generation. If this parameter is set too high, the trees will grow deeper and data will be split more sparsely, leading to trees growing similar and over-fitting of the training data. Based on the experiments, the sampling ratio of $20\%$ and a max\_depth value of $5$ provide the best generalization performance on the glucose monitoring dataset.

Regarding random seeds and feature subsampling, these parameters introduce randomness during random forest generation. The random seeds parameter controls the randomness of the bootstrap of the samples used when generating trees, while feature subsampling ensures that only a proportion of features are available for building trees in one iteration. This prevents the generation of similar or identical trees, which can lead to over-fitting of the meta-train set. Results from Table~\ref{tab:my-table_5} show that randomness is essential in random forest-based models, and its removal significantly decreases the algorithm's accuracy. Therefore, these approaches, including random seeds, feature subsampling, sample sampling, and setting the optimal hyper-parameters, are necessary for improving the generalization performance of random forests and meta-forests.

\subsection{Experiments on Object Recognition Datasets}
In our experiments for object recognition, we used the VLCS and PACS datasets to compare the recognition accuracy of meta-forests with other state-of-the-art models. The VLCS dataset was preprocessed to obtain DeCAF6 features using the DeCAF model by~\cite{cite_18} before inputting the models. On the other hand, the PACS dataset was processed using ResNet-18, as per the setting of other related works. These datasets are commonly used benchmarks for evaluating the performance of various models in the field of object recognition. By conducting experiments on these public datasets, we are able to demonstrate the effectiveness of meta-forests in comparison to other state-of-the-art models.

For both experiments on VLCS and PACS datasets, we conduct $20$ leave-one-domain experiments for every domain. Leave-one-domain experiment is a popular experiment setting for validating model's generalization ability on the unseen domain. We first select one domain dataset as the test set and other domains dataset as the training set. For the next round of experiments, select another domain that was unselected before as the test set, and other domains as the training set. For meta-forests, we split domains in the training set further into meta-train and meta-test datasets as presented in Sections 3 and 4.2. The algorithm will first build the weighted forests model during meta-task iterations before using the target domain (test set) data as the final test. Furthermore, the dataset used for each iteration of meta-forests build is randomly selected $1/5$ of the whole meta-train/test and test set. Therefore, the total data size used for building meta-forests is less than other state-of-art works.

\begin{table}[htb!]
\centering
\caption{Leave-one-domain experiments on object recognition VLCS dataset with state-of-art works.}
\begin{tabular}{@{}cccccc@{}}
\toprule
Model                         & V    & L    & C    & S    & Average (\%) \\ \midrule
CCSA                       &67.1  &62.1  & 92.3 &59.1  & 70.2    \\ 
MMD-AAE                       & 67.7 & 62.6 & 94.4 & 64.4 & 72.3    \\ 
D-MTAE                        & 63.9 & 60.1 & 89.0 & 61.3 & 68.6    \\ 
GCFN                          & 73.8 & 61.7 & 93.9 & 67.5 & 74.2    \\ 
Meta-Reg                      & 65.0 & 60.2 & 92.3 & 64.2 & 70.4    \\ 
MLDG                          & 75.3 &  65.2 &  \textbf{97.4}  & 71.0  &  77.2       \\ \midrule
\textit{\textbf{Meta-forests}} & \textbf{75.5} & \textbf{69.1} & 92.3 & \textbf{76.5} & \textbf{78.4}    \\ \bottomrule
\end{tabular}
\label{tab:my-table_2}
\end{table}

\begin{table}[htb!]
\centering
\caption{Leave-one-domain experiments on object recognition PACS dataset with state-of-art works.}
\begin{tabular}{@{}cccccc@{}}
\toprule
Model       & P    & A     & C     & S     & Average (\%) \\ \midrule
D-MTAE  & 91.1 & 60.3 & 58.7 & 47.9 & 64.5    \\   
Meta-Reg    &  91.1     &  69.8     &  \textbf{70.4}     &  \textbf{59.3}    &    72.7         \\
MLDG        &  88.0     &  66.2     &  66.9     &  59.0     &    70.0          \\\midrule
Meta-Forest & \textbf{91.5} & \textbf{71.4} & 69.7 & 59.2 & \textbf{72.9}       \\ \bottomrule
\end{tabular}
\label{tab:my-table_3}
\end{table}

Tables~\ref{tab:my-table_2} and~\ref{tab:my-table_3} present the results of the leave-one-domain experiments, which demonstrate that our proposed meta-forests outperform all of the state-of-the-art models on almost all of the domains, even the data size is small. These results indicate that meta-forests has strong generalization abilities, not only in processing signal data but also in analyzing image data. However, it is imperative to recognize a limitation in the scalability of meta-forests when compared to model-agnostic methods, such as MLDG. This limitation needs further exploration in future research.

\section{Conclusions}

In this paper, we propose a novel DG algorithm termed meta-forests based on random forests. Our algorithm incorporates the meta-learning strategy and MMD to enhance the generalization ability of the classifiers. Through a series of experiments on two public image recognition datasets and a glucose monitoring dataset collected previously, we have demonstrated that meta-forests outperforms other state-of-the-art approaches in terms of generalization performance. The meta-forests has the potential for the development of ML techniques in areas where data is limited, difficult, or expensive to collect. Furthermore, our work highlights the potential of incorporating meta-learning strategies and regularization techniques in ensemble models. Future research could build on our approach by exploring additional strategies for enhancing the generalization ability of ML models.

\bibliography{acml23}

\end{document}